# Fail-Safe Controller Architectures for Quadcopter with Motor Failures

Gene Patrick S. Rible
*Electrical and Electronics Engineering Institute*
*University of the Philippines*
Quezon City, Philippines
gene.patrick.rible@eee.upd.edu.ph

Nicolette Ann A. Arriola
*Electrical and Electronics Engineering Institute*
*University of the Philippines*
Quezon City, Philippines
nicolette.arriola@eee.upd.edu.ph

Manuel C. Ramos, Jr.
*Electrical and Electronics Engineering Institute*
*University of the Philippines*
Quezon City, Philippines
manuel@eee.upd.edu.ph

*Abstract*—A fail-safe algorithm in case of motor failure was developed, simulated, and tested. For practical fail-safe flight, the quadcopter may fly with only three or two opposing propellers. Altitude for two-propeller architecture was maintained by a PID controller that is independent from the inner and outer controllers. A PID controller on propeller force deviations from equilibrium was augmented to the inner controller of the three-propeller architecture. Both architectures used LQR for the inner attitude controller and a damped second order outer controller that zeroes the error along the horizontal coordinates. The restrictiveness, stability, robustness, and symmetry of these architectures were investigated with respect to their output limits, initial conditions, and controller frequencies. Although the three-propeller architecture allows for distribution of propeller forces, the two-propeller architecture is more efficient, robust, and stable. The two-propeller architecture is also robust to model uncertainties. It was shown that higher yaw rate leads to greater stability when operating in fail-safe mode.

## I. INTRODUCTION

Although stable flight needs at least four propellers, other multicopters employ more for redundancy to achieve safe flight in case of motor failure. Redundant propellers add to the weight and complexity, resulting in greater power consumption and reduction in loop frequency as more propellers are controlled during every iteration.

A solution for quadcopters losing three, two opposing, or one propeller in [1] uses a linear quadratic regulator (LQR) to rotate the quadcopter about a specified axis that is periodically fixed in both inertial and body frame. A 13-camera external motion tracking system is used to continuously provide the quadcopter position and orientation data [2], [3]. Optionally, a pilot sends commands to the quadcopter before and after failure to direct the flight path and land it [4].

This project aims to stabilize and land the quadcopter autonomously after an unexpected motor failure which means that all computations must be performed in flight. The goal is to still achieve fault tolerant return and landing despite limitations in sensor accuracy and computing power.

## II. REVIEW OF RELATED LITERATURE

### A. State Variables and Coordinate Frames

A set of quadcopter coordinate frames $C$ is defined in [5]. The "$x$" superscript in $C^x$ indicates the reference frame. A rotation matrix $R_a^b$ transforms a quantity from $C^a$ to $C^b$. Note that characters in **boldface** are vector or matrix quantities. Table I lists the state variables. The IMU returns $\phi, \theta, \psi, p, q, r$. Meanwhile, $x, y, z, \dot{x}, \dot{y}, \dot{z}$ may be extracted from GPS sensor. $p, q, r$ can be transformed to $\dot{\phi}, \dot{\theta}, \dot{\psi}$ given $\phi, \theta, \psi$ [5].

TABLE I. QUADCOPTER STATE VARIABLES

| | |
|---|---|
| $x$ | inertial position along $\hat{x}^i$ in $C^i$ |
| $y$ | inertial position along $\hat{y}^i$ in $C^i$ |
| $z$ | altitude along $\hat{z}^i$ in $C^i$ |
| $u$ | body frame velocity along $\hat{x}^b$ in $C^b$ |
| $v$ | body frame velocity along $\hat{y}^b$ in $C^b$ |
| $w$ | body frame velocity along $\hat{z}^b$ in $C^b$ |
| $\phi$ | roll angle in $C^{v2}$ |
| $\theta$ | pitch angle in $C^{v1}$ |
| $\psi$ | yaw angle in $C^v$ |
| $p$ | roll angular velocity about $\hat{x}^b$ in $C^b$ |
| $q$ | pitch angular velocity about $\hat{y}^b$ in $C^b$ |
| $r$ | yaw angular velocity about $\hat{z}^b$ in $C^b$ |

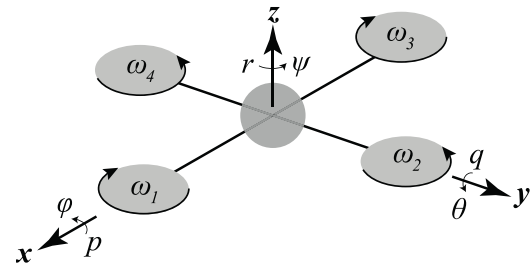

Fig. 1. Quadcopter kinematic diagram. Four propellers are arranged symmetrically about the quadcopter's center of mass. Propellers 1 and 3 rotate opposite to the rotation of propellers 2 and 4. Three coordinate axes *x*, *y*, *z*, are drawn along the quadcopter arms and with origin at the center of mass. The direction of roll, pitch, and yaw angles and their corresponding body frame angular velocity components are indicated by rotation arrows along the coordinate axes.

### B. Dynamic Equations

Equation (1) describes the translational dynamics without air resistance in the inertial frame. Equation (2) describes the rotational dynamics in the body frame. $F = \sum_{i=1}^{4} f_i$ is the total propeller thrust, $M$ is the total quadcopter mass, $J_{zz}^P$ is the

This project was supported in part by grants from the UP Engineering Research & Development Foundation, Inc.

propeller rotational moment of inertia (MOI), $\Omega=\sum_{i=1}^{4}(-1)^i\omega_i$, and the drag $\tau_d=\gamma\, r^2$ where $\gamma$ is the drag coefficient [6].

$$\begin{bmatrix}\ddot{x}\\\ddot{y}\\\ddot{z}\end{bmatrix}=\begin{bmatrix}c\phi s\theta c\psi+s\phi s\psi\\c\phi s\theta s\psi-s\phi c\psi\\c\phi c\theta\end{bmatrix}\frac{F}{M}+\begin{bmatrix}0\\0\\-g\end{bmatrix} \quad (1)$$

$$\begin{bmatrix}\dot{p}\\\dot{q}\\\dot{r}\end{bmatrix}=\begin{bmatrix}\frac{1}{J_{xx}}\tau_\phi\\\frac{1}{J_{yy}}\tau_\theta\\\frac{1}{J_{zz}}\tau_\psi\end{bmatrix}-\begin{bmatrix}\frac{J_{zz}-J_{yy}}{J_{xx}}qr\\\frac{J_{xx}-J_{zz}}{J_{yy}}pr\\\frac{J_{yy}-J_{xx}}{J_{zz}}pq\end{bmatrix}-\begin{bmatrix}\frac{J_{zz}^P}{J_{xx}}q\Omega\\-\frac{J_{zz}^P}{J_{yy}}p\Omega\\\frac{1}{J_{zz}}\tau_d\end{bmatrix} \quad (2)$$

### C. Periodic Equilibrium Solutions for Fail-Safe Flight

In this paper, an overbar ¯ denotes values that are constant along the periodic equilibrium solution, including reference values. A periodic equilibrium solution, as defined in [1], is a certain set of periodic equilibrium values or simply "equilibrium values" that are achieved when the quadcopter has stabilized its flight after a given amount of time from the last change in the translational reference values. The said equilibrium values depend on the quadcopter model parameters and propeller failure condition which is identified herein as the complete failure of three or two opposing motors.

Moreover, $f$ and $\tau$ denote propeller thrust and propeller torque, respectively, with their subscript indices $i$ pertaining to the propeller numbers indicated in Fig. 1.

The strategy adapted in [1] is to control only a single direction of attitude, roll or pitch, once a propeller fails. It is described by primary axis $\bar{n}=(\bar{n}_x,\bar{n}_y,\bar{n}_z)$ that is stationary in $C^i$ and expressed in $C^b$. Without loss of generality, let Motor 4 fail. One possible specification is to force the two opposing propellers to produce equal thrust:

$$\bar{f}_1=\bar{f}_3. \quad (3)$$

A tuning factor may be specified as

$$\rho=\frac{\bar{f}_2}{\bar{f}_1}. \quad (4)$$

Double motor failure solution appears when $\rho=0$. Triple motor failure solution appears as $\rho\to\infty$; although possible in theory, this is impractical as one propeller cannot handle the entire quadcopter's weight.

## III. METHODOLOGY

### A. Hardware

DJI FlameWheel 450 was used as skeleton with E600 propulsion system, 4S Lipo (2800 mAh), YOST 3-Space IMU, Ublox NEO-M8N GPS, HC-SR04 ultrasonic sensor, and Raspberry Pi Zero W. Each motor was connected to a wireless relay which may be remotely activated during flight to induce failure. DJI E600 can receive motor signals between 30 Hz and 450 Hz, limiting the achievable effective loop frequencies.

### B. Modeling

*1) Axial MOIs:* To obtain the axial MOIs, the quadcopter was suspended on three different axes and made to swing freely at small angles. The MOI about the axis at the pivot point is

$$J_P=Mgr\left(\frac{T}{2\pi}\right)^2 \quad (5)$$

where $T$ is the period of oscillation and $r$ is the distance between the center of mass (COM) and pivot point [7]. Using parallel axis theorem, the desired axial MOIs can be obtained. Meanwhile, MOI of propellers were obtained by assuming that the motor is a solid cylinder and the propeller blade a circular disk with evenly distributed masses.

*2) Propeller Characterization:* The thrust magnitude, torque magnitude, and angular speed of a rotating propeller blade are related by [8]

$$f_i=k_f\omega_i^2 \quad (6)$$

$$\tau_i=k_\tau\omega_i^2 \quad (7)$$

$$f_i=k\tau_i. \quad (8)$$

Note that $k_f$, $k_\tau$, and $k$ are model parameters to be experimentally determined. In characterizing the propellers, the microcontroller sends a PWM command to the propeller. As the propeller spins, a weighing scale measures the thrust while a tachometer measures the propeller angular speed. The reaction torque is computed from the voltage and current drawn. At steady state, electrical torque $\tau_e$ equals mechanical torque $\tau_m$ [9]. The propeller exerts a torque in a direction opposite its sense of rotation [10]—a part of this is the reaction torque $\tau_\psi$ exerted by the spinning propeller onto the body frame whereas the remaining is dissipated by aerodynamic drag $\tau_{d_{prop}}$. Thus, at steady state,

$$\tau_e=\tau_m=\tau_\psi+\tau_{d_{prop}}. \quad (9)$$

From (7),

$$\tau_\psi=k_\tau\omega^2 \quad (10)$$

where $k_\tau$ is a positive scalar. It was also shown in [11] that

$$\tau_{d_{prop}}=k_d\omega^2 \quad (11)$$

where $k_d$ is also a positive scalar. Since $\tau_\psi$ and $\tau_{d_{prop}}$ behave similarly with respect to $\omega$, it is impossible to distinguish between the two. We can set $\tau_{d_{prop}}=0$ so that

$$\tau_\psi=\tau_e-\tau_{d_{prop}}=\tau_e. \quad (12)$$

$\tau_{d_{prop}}$ is lumped together with the drag torque $\tau_d$ on the body frame via experimental modeling. This effectively eliminates whatever excess torque allotment that went into $\tau_\psi$.

*3) Drag Coefficient:* In Fig. 2, the quadcopter is attached to a freely rotatable bearing. Commanding two opposing propellers to exert same torque rotates the quadcopter about $\hat{z}^b$. Each torque value corresponds to a steady state angular velocity. Parametrizing the plot of the total torque versus the steady state angular velocity yields the drag coefficient.

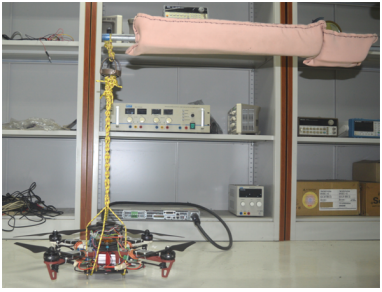

Fig. 2. Rope and bearing setup. The bearing allows the quadcopter in flight to rotate freely along the yaw direction.

### C. Motor Failure Detection and Identification

Each combination of failure(s) may be uniquely identified by spikes in state readings. Table II summarizes all possible cases. Note that a propeller exerts a torque opposite its sense of rotation onto the quadcopter.

TABLE II. SIGNATURE SPIKES OF MOTOR FAILURE COMBINATIONS

| Motor Failure | $p$ | $q$ | $r$ | $\phi$ | $\theta$ |
|---|---|---|---|---|---|
| 1 | | ↑ | ↓ | | ↑ |
| 2 | ↓ | | ↑ | ↓ | |
| 3 | | ↓ | ↓ | | ↓ |
| 4 | ↑ | | ↑ | ↑ | |
| 1, 3 | | | ↓ | | |
| 2, 4 | | | ↑ | | |
| 2, 3, 4 | ↓ | ↑ | | ↓ | |
| 1, 3, 4 | ↑ | | | ↑ | |
| 1, 2, 4 | | ↑ | ↑ | | ↑ |
| 1, 2, 3 | ↓ | | ↓ | ↓ | |

### D. Controller Architecture

*1) Inner Controller:* The state vector is $s=(p,q,n_x,n_y)$ with state error $\tilde{s}=s-\bar{s}$ which evolves to first order as $\dot{\tilde{s}}=A\tilde{s}+Bu$ where, based on (2),

$$A=\frac{\partial \dot{\tilde{s}}}{\partial \tilde{s}}=-\begin{bmatrix} 0 & \bar{a} & 0 & 0 \\ \bar{b} & 0 & 0 & 0 \\ 0 & \bar{n}_z & 0 & -\bar{r} \\ -\bar{n}_z & 0 & \bar{r} & 0 \end{bmatrix} \quad (13)$$

with

$$\bar{a}=\frac{J_{zz}-J_{yy}}{J_{xx}}\bar{r}+\frac{J_{zz}^P}{J_{xx}}(-\bar{\omega}_1+\bar{\omega}_2-\bar{\omega}_3+\bar{\omega}_4) \quad (14)$$

$$\bar{b}=\frac{J_{xx}-J_{zz}}{J_{yy}}\bar{r}-\frac{J_{zz}^P}{J_{yy}}(-\bar{\omega}_1+\bar{\omega}_2-\bar{\omega}_3+\bar{\omega}_4). \quad (15)$$

System input is introduced as a function of propeller force deviations from equilibrium by specifying

$$B=\begin{bmatrix} 0 & \frac{l}{J_{xx}} & 0 \\ -\frac{l}{J_{yy}} & 0 & \frac{l}{J_{yy}} \\ 0 & 0 & 0 \\ 0 & 0 & 0 \end{bmatrix} \quad (16)$$

with $u=(u_1,u_2,u_3)$ and $u_i=f_i-\bar{f}_i$.

*a) Three-Propeller:* Additional PID controller on propeller force deviations from equilibrium was added so that

$$f_i=u_i+\bar{f}_i+u_{\tilde{f}_i} \quad (17)$$

where $\tilde{f}_i=f_i-\bar{f}_i$ and $u_{\tilde{f}_i}$ is the PID output.

*b) Two-Propeller:* With two propellers, altitude PID control can be decoupled from the LQR controller because of symmetry, resulting in a more robust and stable system. The inner controller is described by

$$f_i=u_i+\bar{f}_i+u_z \quad (18)$$

where $u_z$ is the PID output. Taking into account the tilt which reduces lift force, $u_z$ is divided by $\hat{z}^b\cdot\hat{z}^i=\cos(\phi)\cos(\theta)$.

*2) Outer Controller:* $\tilde{d}=(\tilde{x},\tilde{y},\tilde{z})$ is the translational deviation of the quadcopter from a desired point in space. As employed in [1], the outer controller zeroes out position error according to damping ratio $\zeta$ and natural frequency $\omega_n$ by introducing a desired translational deviation acceleration $\ddot{\tilde{d}}_{des}$ such that

$$\ddot{\tilde{d}}_{des}+2\zeta\omega_n\dot{\tilde{d}}+\omega_n^2\tilde{d}=0 \quad (19)$$

where $\zeta$ and $\omega_n$ are diagonal matrices. The instantaneously desired direction of $n$ is given by [1]

$$R_b^v n_{des}\bar{n}_z F+Mg=M\ddot{d}_{des}. \quad (20)$$

In (20), the component of $n$ along $\hat{z}^b$ is shifted such that the total periodic thrust and gravity act on the COM to achieve the desired acceleration. For two-propeller architecture, $\zeta_{zz}$ and $\omega_{n_{zz}}$ were set to 0 so that the quadcopter's altitude is maintained solely by the PID controller. The implemented frequency of the inner controller was 450 Hz while the outer translational controller (and altitude controller as well for the two-propeller case) interrupted at 45 Hz. However, the achievable effective translational controller frequency was only 10 Hz to match the GPS update rate.

## E. Sensor Fusion and Filters

For attitude sensing, accelerometer and magnetometer estimates were fused with gyroscope estimates via complementary filtering. Gyroscope readings were pre-filtered with exponential moving average filter. The tuning of filter time constants was done in powers of 2 of the loop period $\frac{1}{450\text{ Hz}}$.

## IV. RESULTS

Table III shows the quadcopter parameters. The high inertia model is used in Section IV-H; it corresponds to a physical quadcopter modeled using the methods presented.

TABLE III.  QUADCOPTER PARAMETERS

| Parameter | Low Inertia | High Inertia |
|---|---|---|
| $M$ | 1.439 kg | 1.988 kg |
| $l$ | 0.2475 m | 0.2475 m |
| $g$ | 9.80665 $\frac{m}{s^2}$ | 9.80665 $\frac{m}{s^2}$ |
| $J_{xx}$ | 0.018517242 kg m$^2$ | 0.125203794 kg m$^2$ |
| $J_{yy}$ | 0.020562251 kg m$^2$ | 0.120414017 kg m$^2$ |
| $J_{zz}$ | 0.028316170 kg m$^2$ | 0.163195234 kg m$^2$ |
| $J_{zz}^P$ | 9.76065E–05 kg m$^2$ | 2.66838E–04 kg m$^2$ |
| $\gamma$ | 0.000184199 $\frac{\text{N m}}{(\text{rad}/\text{s})^2}$ | 0.00258396780706647 $\frac{\text{N m}}{(\text{rad}/\text{s})^2}$ |

In the following figures, blue plots correspond to actual values; red plots to equilibrium or reference values; green plots to updated reference values.

### A. Three-Propeller Simulation

With $\rho=0.5$, Fig. 3 and 4 show the simulated three-propeller quadcopter response. The tuned nonzero gains are $Q_{p,q}=1\left(\frac{\text{rad}}{\text{s}}\right)^{-2}$, $Q_{n_{x,y}}=20$, $R_{f_1}=1.11\text{ N}^{-2}$, $R_{f_2}=10\text{ N}^{-2}$, $R_{f_3}=1\text{ N}^{-2}$, $k_{i,f1}=0.129\text{ s}^{-1}$, $k_{i,f2}=0.05\text{ s}^{-1}$, $k_{i,f3}=0.114\text{ s}^{-1}$, $\omega_{n,(x,y)}=1\frac{\text{rad}}{\text{s}}$, $\zeta_{x,y}=0.7$, $\omega_{n,z}=2.1\frac{\text{rad}}{\text{s}}$, $\zeta_z=4.5$.

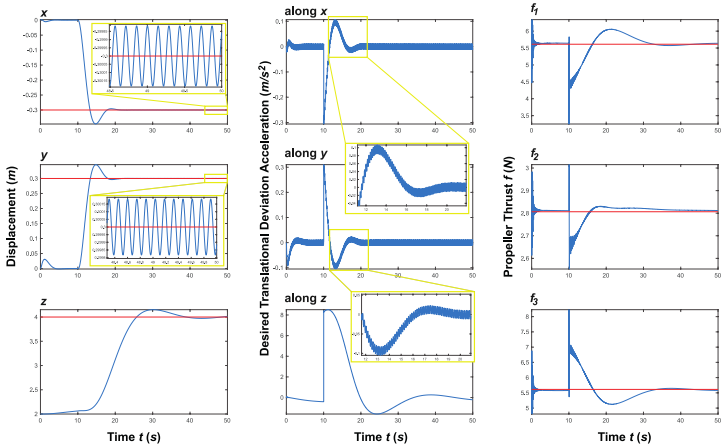

Fig. 3. Simulated response of three-propeller architecture – part 1. The first column shows the quadcopter translational coordinates. The second column shows the computed desired translational deviation acceleration components needed to achieve stabilization to reference translational values. The third column shows the thrusts needed from each propeller to achieve the desired translational deviation acceleration.

The equilibrium values are $\bar{q}=1.50780\frac{\text{rad}}{\text{s}}$, $\bar{r}=43.3930\frac{\text{rad}}{\text{s}}$, $\bar{n}_y=0.034727$, $\bar{n}_z=0.9994$, $\bar{f}_{1,3}=5.6128\text{ N}$, $\bar{f}_2=2.80640\text{ N}$, $\bar{\omega}_{1,3}=710.67829\frac{\text{rad}}{\text{s}}$, $\bar{\omega}_{1,3}=487.1272\frac{\text{rad}}{\text{s}}$, and $\epsilon=0.023031$.

In this simulation, the quadcopter with only three propellers active is initially at equilibrium when the translational reference values are changed at 10 s to $x=-0.3$ m, $y=0.3$ m, $z=4$ m. It can be seen from Fig. 3 that as the quadcopter moves toward these translational reference values as shown in the first column, the required propeller thrusts shown in the third column steadily approach equilibrium and the desired translational deviation acceleration components shown in the second column approach zero.

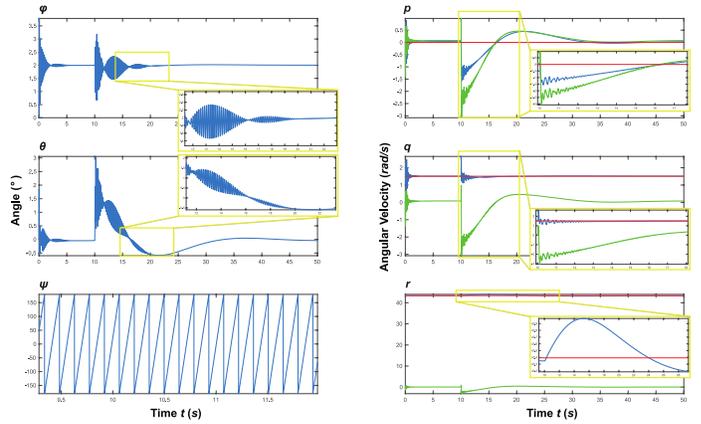

Fig. 4. Simulated response of three-propeller architecture – part 2. The first column shows the quadcopter's roll, pitch, and yaw. The second column shows its body frame angular velocity components.

Meanwhile, Fig. 4 shows the quadcopter's orientation as well as its body frame angular velocities which, as discussed in Section IV-G, form the basis for the primary axis. Notice that although there is no static equilibrium for the roll, pitch, and yaw angles, the angular velocity components approach equilibrium values at steady state. Note that the zigzag pattern of the yaw plot indicates the quadcopter's continuous, rapid spinning motion along the yaw direction.

### B. Two-Propeller Simulation and Implementation

Fig. 5 and 6 show the simulated two-propeller response. To initiate the controller, the quadcopter is initially placed at a small displacement (i.e. at coordinates $x=-0.1$ m, $y=-0.1$ m, $z=2$ m) from initial reference values $x=0$ m, $y=0$ m, $z=2$ m.

With the rope and bearing setup in Fig. 2, it was possible to safely test the attitude and altitude controller but not the outer translational controller. In the implementation, the quadcopter will be tasked to stabilize about a vertical primary axis that is pointing opposite to gravity. The three-propeller case could not be safely tested with the current setup in the lab due to the circular trajectory [1].

Fig. 7 shows the implemented two-propeller response. Note from Fig. 8 that the quadcopter was able to maintain an altitude close to a reference 0.7 m above ground. Because the propellers were operating very close to limits, maximum PID output had to be capped at 5.0 N to still give a room for attitude controller. Meanwhile, minimum PID output was capped at –0.5 N to prevent quick descent. The strategy employed was to increase the proportional gain while relying on the PID output capping limits so that the PID output becomes similar to a bang-bang controller as seen in Fig. 9. Table IV lists the tuned controller gains. Even though the simulated and implemented optimal gains are far apart due to unmodeled factors, when the implemented gains are simulated, they work to stabilize the system. An accompanying video in [12] demonstrates the controller gains tuning process.

Nevertheless, the angular velocity components in Fig. 7 are shown to stabilize within boundaries and in the actual implementation, the quadcopter was physically able to stabilize about the vertical primary axis as demonstrated in [12].

TABLE IV. CONTROLLER VALUES FOR TWO-PROPELLER ARCHITECTURE

| Parameter | Simulated | Implemented |
| --- | --- | --- |
| $Q_p$ | $0 \left(\frac{\text{rad}}{\text{s}}\right)^{-2}$ | $0 \left(\frac{\text{rad}}{\text{s}}\right)^{-2}$ |
| $Q_q$ | $0 \left(\frac{\text{rad}}{\text{s}}\right)^{-2}$ | $0 \left(\frac{\text{rad}}{\text{s}}\right)^{-2}$ |
| $Q_{n_x}$ | 5362 | 42 |
| $Q_{n_y}$ | 5362 | 1.9 |
| $R_{f_1}$ | $1\ \text{N}^{-2}$ | $0.9\ \text{N}^{-2}$ |
| $R_{f_3}$ | $1\ \text{N}^{-2}$ | $0.9\ \text{N}^{-2}$ |
| $k_{p,z}$ | $4.3\ \frac{\text{N}}{\text{m}}$ | $75\ \frac{\text{N}}{\text{m}}$ |
| $k_{d,z}$ | $8.9\ \frac{\text{N s}}{\text{m}}$ | $3\ \frac{\text{N s}}{\text{m}}$ |
| $k_{i,z}$ | $0\ \frac{\text{N}}{\text{m s}}$ | $3\ \frac{\text{N}}{\text{m s}}$ |
| $\omega_{n,x}$ | $1\ \frac{\text{rad}}{\text{s}}$ | disabled |
| $\zeta_x$ | 0.7 | disabled |
| $\omega_{n,y}$ | $1\ \frac{\text{rad}}{\text{s}}$ | disabled |
| $\zeta_y$ | 0.7 | disabled |

### C. Horizontal Circular Trajectory

From Fig. 3, x, y are oscillating at an amplitude of 0.3001 m–0.2999 m=0.0002 m, forming a circular trajectory with radius $\sqrt{\left(\frac{0.0002\ \text{m}}{2}\right)^2 + \left(\frac{0.0002\ \text{m}}{2}\right)^2}$ =0.0001414 m , close to the theoretical value. With two propellers, no circular trajectory is formed so Fig. 5 shows no oscillations along *x*, *y* which is more efficient as thrusts are fully used for lift instead of revolving the rotating quadcopter unnecessarily. Nevertheless, three-propeller architecture allows for distribution of thrusts for fail-safe flight thereby lowering requirements from each propeller.

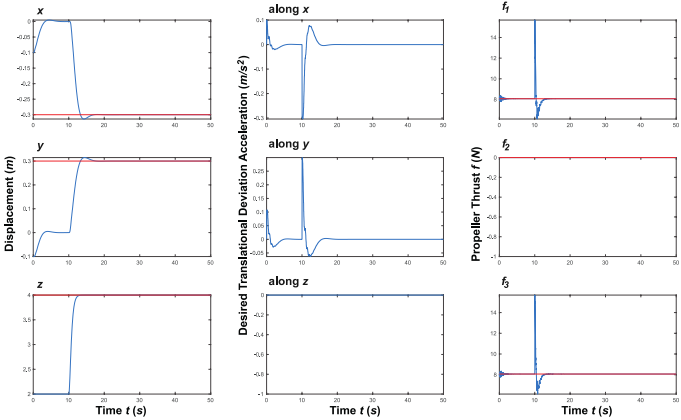

Fig. 5. Simulated response of the two-propeller architecture – part 1. The translational reference values, initially *x*=0 m, *y*=0 m, *z*=2 m, are changed at 10 s to *x*=–0.3 m, *y*=0.3 m, *z*=4 m.

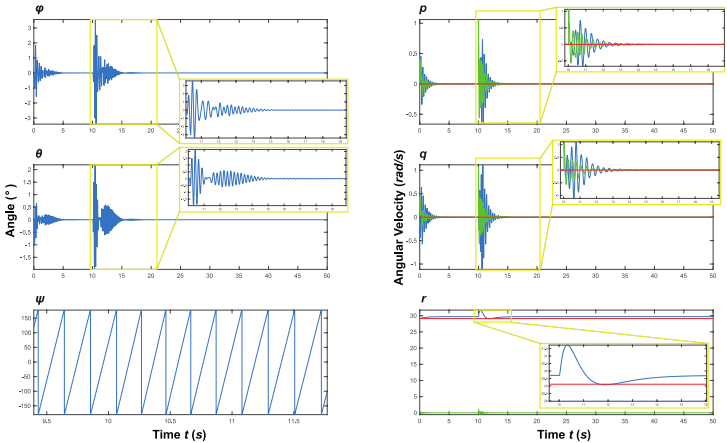

Fig. 6. Simulated response of the two-propeller architecture – part 2. The equilibrium values are $\bar{r}$=32.021 $\frac{\text{rad}}{\text{s}}$, $\bar{f}_{1,3}$=7.0559 N, $\bar{\omega}_{1,3}$=803.9458 $\frac{\text{rad}}{\text{s}}$.

Observe in Fig. 7 that the orientation angles and angular velocity measurements in the implementation are noisier and yield wider oscillations than the simulation, as shown in Fig. 6, predicts. This discrepancy is mostly due to sensor noise and a lot of unmodeled non-idealities in the implementation.

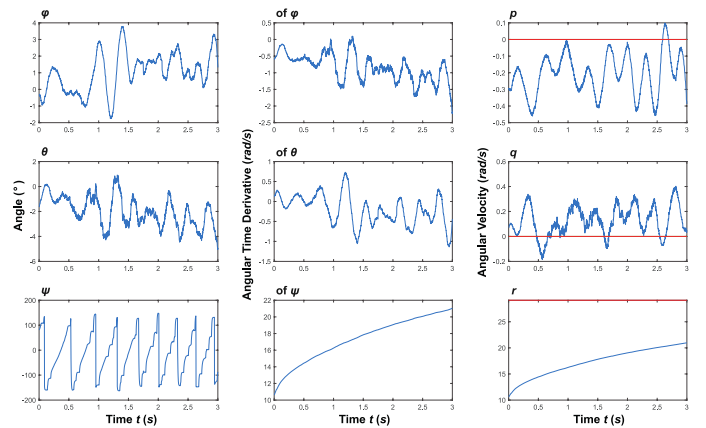

Fig. 7. Implemented two-propeller attitude response while hovering. The quadcopter's roll, pitch, and yaw are shown in the first column and their time derivatives are shown in the second column. The third column shows the quadcopter's body frame angular velocity components.

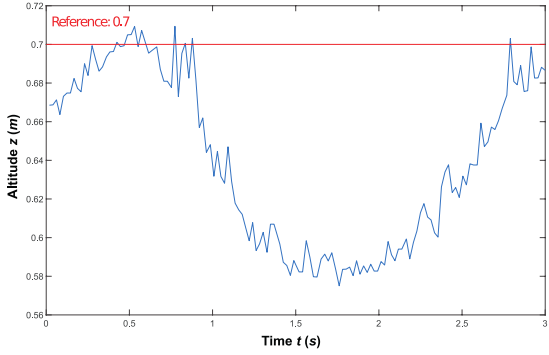

Fig. 8. Implemented two-propeller altitude response while hovering. Using the altitude PID controller, the quadcopter is trying to maintain its altitude (in blue) close to the reference altitude (in red).

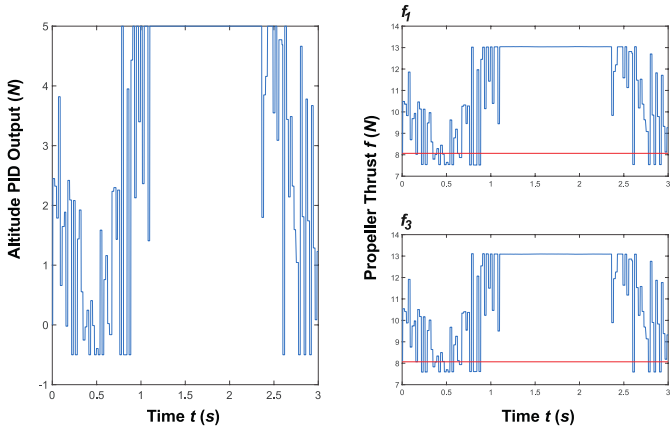

Fig. 9. Implemented two-propeller altitude PID output and propeller thrusts while hovering. The first column shows the PID output that must be added to the thrust command values of propellers 1 and 3 to maintain reference altitude. The second column shows the actual thrust values commanded to the propellers in order to maintain reference altitude while achieving flight stabilization. Note that the change in PID output values happens in discrete intervals, at a lower frequency than the quadcopter's inner attitude controller.

### D. Output Limits

The three-propeller architecture works for any change along $x$, $y$ if $\ddot{x}_{des}$, $\ddot{y}_{des}$ are capped at 2.1 $\frac{m}{s^2}$ in magnitude. It also works for any change along $z$ if $\ddot{z}_{des}$ is capped at 16.5 $\frac{m}{s^2}$. Meanwhile, the two-propeller architecture $\ddot{x}_{des}$, $\ddot{y}_{des}$ must be capped at 5.4 $\frac{m}{s^2}$. Its altitude PID output must be capped at 1.1 N. These simulated limits are greater than the three-propeller, indicating greater stability.

### E. Initial Condition Limits

The three-propeller architecture would still work if state variables deviate from equilibrium within the limits in Table V. With Motor 4 failure, symmetry in $\phi$ and $p$ is lost so the positive and negative limits in $\theta$ and $q$ are more symmetric. There is no limit to the initial values of $\dot{x}$, $\dot{y}$; the system would eventually stabilize but the larger the magnitudes of $\dot{x}$ and/or $\dot{y}$, the longer it would take the system to settle down to near-zero speeds. Moreover, the quadcopter would have to build up angular momentum until $r$ exceeds 43.3930 $\frac{rad}{s}$ –14.8 $\frac{rad}{s}$ =28.6 $\frac{rad}{s}$, assuming all other variables are initially at equilibrium.

Meanwhile, two-propeller limits are generally way higher, indicating greater stability. Except for translational variables, positive and negative limits are symmetric. The system can theoretically stabilize even for initial $\phi$, $\theta$ beyond ±90°. The required initial $r$ is 32.021 $\frac{rad}{s}$ –6.6 $\frac{rad}{s}$ =25.4 $\frac{rad}{s}$ which is lower.

### F. Frequency Limits

From simulations, frequency of outer three-propeller translational plus altitude controller must be >17 Hz whereas inner attitude controller must be >60 Hz. For the two-propeller architecture, outer controller must be >11 Hz whereas inner controller must be >14 Hz. The lower frequency requirements indicates greater robustness and stability. As the ratio of frequencies approach unity, the minimum frequency for the outer controller becomes slightly more restrictive.

TABLE V. FAIL-SAFE CONTROLLER INITIAL CONDITION LIMITS

| State Variable | Three-Propeller | | Two-Propeller | |
|---|---|---|---|---|
| | + | − | + | − |
| $\phi$ | 22.2° | −16.2° | 109.7° | −111.0° |
| $\theta$ | 20.0° | −20.1° | 103.5° | −104.0° |
| $p$ | 19.9 $\frac{rad}{s}$ | −24.1 $\frac{rad}{s}$ | 19.2 $\frac{rad}{s}$ | −18.9 $\frac{rad}{s}$ |
| $q$ | 16.2 $\frac{rad}{s}$ | −19.0 $\frac{rad}{s}$ | 25.2 $\frac{rad}{s}$ | −25.5 $\frac{rad}{s}$ |
| $r$ | 212.7 $\frac{rad}{s}$ | −14.8 $\frac{rad}{s}$ | ∞ | −6.6 $\frac{rad}{s}$ |
| $x$ | ∞ | −∞ | ∞ | −∞ |
| $y$ | ∞ | −∞ | ∞ | −∞ |
| $z$ | ∞ | −∞ | ∞ | −∞ |
| $\dot{x}$ | ∞ | −∞ | ∞ | −∞ |
| $\dot{y}$ | ∞ | −∞ | ∞ | −∞ |
| $\dot{z}$ | ∞ | −0.5 $\frac{m}{s}$ | 11.0 $\frac{m}{s}$ | −∞ |

### G. Effect of r on Fail-Safe Stability

From periodic solution equations [1], it follows that

$$n_x = \frac{p}{\sqrt{p^2+q^2+r^2}}. \qquad (21)$$

$n_y$, $n_z$ are similarly obtained with $q$, $r$ in the numerator. When $r$ increases with $p$, $q$ constant, $n_z$ increases and approaches 1 while $n_x$, $n_y$ approach 0. With three propellers, either $\bar{n}_x$ or $\bar{n}_y$ is 0; with two propellers, both $\bar{n}_x$ and $\bar{n}_y$ are 0. Increasing $r$ by decreasing $\tau_d$ or $J_{zz}$ unconditionally lowers $n_x$, $n_y$ to their equilibrium values, thereby helping to stabilize the system. Furthermore, (22) and (23) suggest that that there is more average thrust along $\hat{z}^i$ if $\bar{n}_z$ is higher so requirements from each propeller are generally lower. Note that the horizontal circular trajectory radius $\bar{R}_{ps}$ decreases as $\bar{n}_z$ increases, resulting in greater efficiency. These observations were confirmed by simulating and experimentally testing a quadcopter with higher inertia and drag.

$$\bar{F}\bar{n}_z = Mg \qquad (22)$$

$$\bar{R}_{ps} = \frac{\sqrt{1-\bar{n}_z^2}}{\bar{n}_z} \frac{g}{\|\bar{\omega}\|^2} \tag{23}$$

where $\bar{\omega}$ is the quadcopter's equilibrium body frame angular velocity.

*H. Robustness of Fail-Safe Controller to Model Uncertainty*

When flying in fail-safe with two propellers so that $\bar{p}=\bar{q}=0$, the resulting $\bar{r}$ and $\epsilon$ which depend only on effective drag coefficients are reliable. Drag coefficients will always yield a value for $\bar{r}$ in accordance with the experimentally extracted model between the assumed propeller torque and observed steady state $r$. Thus, lumped in the coefficients are all uncertainties in $J_{zz}$ as well as modeling inaccuracies of propellers. This means that although the system's model may be uncertain, the resulting equilibrium values when flying with two propellers are reliable using the modeling methods discussed in this paper.

Simulations show that although the assumed system used to calculate the LQR gains is not the true system, it is possible to find a set of gains that can stabilize the true system, provided that the equilibrium values are accurate enough. A different quadcopter with higher inertia was successfully stabilized in simulations using the low inertia model in Table III. The tuned nonzero gains are $Q_{n_{x,y}}=420$, $R_{f_{1,3}}=1\text{ N}^{-2}$, $k_{p,z}=2.8\ \frac{\text{N}}{\text{m}}$, $k_{d,z}=3.2\ \frac{\text{N s}}{\text{m}}$, $\omega_{n,(x,y)}=1\ \frac{\text{rad}}{\text{s}}$, $\zeta_{x,y}=0.7$.

## V. Conclusions

Fail-safe algorithm in case of motor failure for autonomous quadcopter with only onboard sensors and processor was developed, simulated, and tested. This paper also presents a method to detect the failure of a motor or particular combinations of motors. Since system model is vital to the controller, this paper additionally presents methods to model the quadcopter without needing expensive and least available apparatuses such as a reaction torque sensor.

In theory, there can be one-propeller fail-safe architecture but this is impractical as only one cannot handle the quadcopter's weight. Although three-propeller architecture allows for distribution of thrusts thereby lowering propeller requirements, two-propeller architecture is more efficient. Moreover, decoupling of altitude from translational and attitude control makes the two-propeller architecture more robust and stable. Although simulated frequency limits vary depending on quadcopter parameters and controller gains, lower limit for the outer translational controller was usually below 20 Hz. Although the GPS receiver used in this project had an update rate of around 10 Hz only, modern GPS receivers are capable of 5 Hz to 20 Hz, depending on location [13]. Additionally, simulated inner controller frequency limit was usually below 20 Hz for two-propeller architecture and below 100 Hz for three-propeller architecture. Most modern onboard attitude sensors and processors are highly capable of achieving these. It was shown that higher yaw rate leads to improved efficiency and greater stability in fail-safe mode, meaning larger and heavier unmanned aerial vehicles or those with greater air resistance coefficients are harder and more inefficient to stabilize. Finally, the two-propeller architecture is robust to model uncertainties so even if the model used is not true, it is possible to find controller gains that stabilize the system using the uncertain model. The extent of uncertainty allowed and other requirements may be explored in another study.